\documentclass[11pt]{article}

\usepackage{array}

\newcolumntype{M}[1]{>{\centering\arraybackslash}m{#1}}

\usepackage{booktabs} 

\usepackage{hyperref}

\usepackage{color,soul}
\usepackage{xcolor}
\usepackage{framed}
\colorlet{shadecolor}{yellow!20}

\usepackage{tikz}
\usepackage{verbatim}

\usepackage{amssymb}
\usepackage{import}
\usepackage{lineno}
\usepackage{mathtools,bbm}
\usepackage{graphicx,url}
\usepackage{graphics}
\usepackage[a4paper, margin=1in]{geometry}
\usepackage{pdflscape}
\usepackage{tabularx} 

\usepackage{listings}
\DeclareUnicodeCharacter{2265}{\ensuremath{\geq}}

\usepackage{setspace,graphicx,epstopdf,amsmath,amsfonts,amssymb,amsthm,versionPO}
\usepackage{marginnote,datetime,enumitem,rotating,fancyvrb,scalerel}
\usepackage{hyperref}

\excludeversion{notes}		
\includeversion{links}          

\iflinks{}{\hypersetup{draft=true}}

\ifnotes{%
\usepackage[margin=1in,paperwidth=10in,right=2.5in]{geometry}%
\usepackage[textwidth=1.4in,shadow,colorinlistoftodos]{todonotes}%
}{%
\usepackage[margin=1in]{geometry}%
\usepackage[disable]{todonotes}%
}

\makeatletter\let\chapter\@undefined\makeatother 

\usepackage{indentfirst} 
\usepackage{jfe}          

\usepackage[
  backend=biber,natbib,
  sorting=none, 
   style=authoryear,
]{biblatex}
\DefineBibliographyStrings{german}{%
  andothers = {et al.},
}

\usepackage{float}
\usepackage[caption = false]{subfig}
\usepackage{blindtext}
\usepackage{multirow}
\usepackage{bm}

\begin{document}

\setlist{noitemsep}  

\title{\textbf{Generative AI for Urban Planning: Synthesizing Satellite Imagery via Diffusion Models}}


\author{
   \small{Qingyi Wang\textsuperscript{1}, Yuebing Liang\textsuperscript{3}, Yunhan Zheng\textsuperscript{3}, Kaiyuan Xu\textsuperscript{4}, Jinhua Zhao\textsuperscript{5}, Shenhao Wang\textsuperscript{2,3}\textsuperscript{*}} \\
  \\
  \small{1 - Department of Civil and Environment Engineering, Massachusetts Institute of Technology} \\
  \small{2 - Department of Urban and Regional Planning, University of Florida} \\
  \small{3 - The Singapore-MIT Alliance for Research and Technology} \\
  \small{4 - Department of Systems Engineering, Boston University} \\
  \small{5 - Department of Urban Studies and Planning, Massachusetts Institute of Technology} 
}



\renewcommand{\thefootnote}{\fnsymbol{footnote}}
\renewcommand{\figurename}{Figure}
\singlespacing
\maketitle

\vspace{-.2in}
\begin{abstract}
\noindent
Generative AI offers new opportunities for automating urban planning by creating site-specific urban layouts and enabling flexible design exploration. However, existing approaches often struggle to produce realistic and practical designs at scale. Therefore, we adapt a state-of-the-art Stable Diffusion model, extended with ControlNet, to generate high-fidelity satellite imagery conditioned on land use descriptions, infrastructure, and natural environments. To overcome data availability limitations, we spatially link satellite imagery with structured land use and constraint information from OpenStreetMap. Using data from three major U.S. cities, we demonstrate that the proposed diffusion model generates realistic and diverse urban landscapes by varying land-use configurations, road networks, and water bodies, facilitating cross-city learning and design diversity. We also systematically evaluate the impacts of varying language prompts and control imagery on the quality of satellite imagery generation. Our model achieves high FID and KID scores and demonstrates robustness across diverse urban contexts. Qualitative assessments from urban planners and the general public show that generated images align closely with design descriptions and constraints, and are often preferred over real images. This work establishes a benchmark for controlled urban imagery generation and highlights the potential of generative AI as a tool for enhancing planning workflows and public engagement. \\

\noindent Corresponding to: Shenhao Wang - \url{shenhaowang@ufl.edu}
\end{abstract}

\medskip


\thispagestyle{empty}

\clearpage

\onehalfspacing
\setcounter{footnote}{0}
\renewcommand{\thefootnote}{\arabic{footnote}}
\setcounter{page}{1}

\section{Introduction}
\label{sec:4-intro}
\noindent

\noindent Urban planning is a complex, iterative, and resource-intensive process, in which visualization has been used to facilitate decision-making at each stage.
Urban planners first articulate projects' objectives and assess the neighborhood's existing infrastructure, natural environment, and sociodemographics. Urban planners will then design urban landscapes to achieve the goals and preserve certain components in the surrounding infrastructure and natural environment.
The plans are then iterated over feedback collected from a multitude of stakeholders, including local government, community residents, and real estate developers. Effective communication among these stakeholders is paramount, and it is particularly crucial to engage the public through visualized urban landscapes, which provide intuitive perspectives to the non-experts \citep{Kempenaar_Westerink_van_Lierop_Brinkhuijsen_van_den_Brink_2016, Mueller_Lu_Chirkin_Klein_Schmitt_2018}. 
In such a process, quick and realistic visualizations are pivotal in bridging gaps in understanding, fostering effective communication, and achieving stakeholder consensus.

Generative Artificial Intelligence (GenAI) presents an opportunity to significantly expedite the process of planning, communication, and feedback. Recently, AI technologies have exhibited exciting potential in various aspects of urban planning, including gaining data-driven insights, evaluating and optimizing performance, and creating visualizations. 
One distinct advantage AI has over humans is the ability to learn from large amounts of data. AI models can produce deep insights, identify trends and patterns in complex city ecosystems, and generate optimal land use and building layouts \citep{Wang_Fu_Liu_Chen_Wang_Lu_2023, Wang_Wu_Zhang_Zhou_Sun_Fu_2023, Zheng_Lin_Zhao_Wu_Jin_Li_2023}. The insights help human planners make more informed and data-driven decisions. Additionally, AI can help evaluate and optimize the performance of urban plans. For example, Delve\footnote{\url{https://www.sidewalklabs.com/products/delve}} from Sidewalk Labs and Forma\footnote{\url{https://www.autodesk.com/products/forma/overview}} from Autodesk are two commercial tools capable of generating and evaluating designs. 
Lastly, GenAI brings more potential to learn and apply visual styles for quick visualization of completed plans with image-to-image models \citep{Ye_Du_Ye_2022, Espinosa_Crowley_2023}. 

Although GenAI can potentially transform the urban planning process, at least three challenges remain. First, current research often falls short in addressing the diverse and complex conditions inherent in urban planning, including site-specific constraints and design descriptions from human experts. Site constraints include existing infrastructure and natural environment, while design descriptions can encompass land use proportions, road and building density, and other detailed specifications. Second, existing GenAI-based urban planning solutions have primarily relied on generative adversarial networks (GANs) \citep{Wang_Fu_Liu_Chen_Wang_Lu_2023, Wang_Wu_Zhang_Zhou_Sun_Fu_2023, Zheng_Lin_Zhao_Wu_Jin_Li_2023}, which often struggle to produce high-quality imagery due to issues like mode collapse, training instability, and poor scalability to larger architectures and datasets \citep{Croitoru_Hondru_Ionescu_Shah_2023}. Recently, diffusion models have emerged as a more robust and stable alternative, consistently generating high-quality imagery across various domains \citep{Ho_Jain_Abbeel_2020, Dhariwal_Nichol_2021, Croitoru_Hondru_Ionescu_Shah_2023}. Lastly, effective training of diffusion models requires large amounts of data, and in the context of urban planning, labeled data is often expensive and relatively scarce, limiting the applicability of the diffusion models. 

This work addresses the challenges by extending the stable diffusion model, leveraging widely available OpenStreetMap data to generate large-scale urban landscapes represented by satellite imagery. The stable diffusion model facilitates model training and improves upon the GAN models. The open-access OpenStreetMap data enables us to expand our research scope to three major metropolitan areas. 

Overall, this work makes the following five contributions\footnote{To promote open science, our scripts and data processing can be found in the repository at \url{https://github.com/sunnyqywang/Urban-Control}.}
\begin{enumerate}
    \item We developed a generative urban planning framework that can automatically generate urban landscapes based on site-specific constraints and design descriptions. This framework demonstrates the potential of generative AI as a powerful visualization tool for automating the urban planning process. 
    \item We adapted a state-of-the-art diffusion model to generate high-fidelity, realistic satellite imagery corresponding to land use descriptions, existing infrastructure, and natural environments across various urban contexts. 
    \item We proposed a data processing pipeline based on open-source and globally available OpenStreetMap and satellite imagery, offering a solution to the challenge of scarce labeled data in the urban setting. 
    \item We used FID and KID scores to measure fidelity across imagery controls and textual prompts, thus establishing a benchmark for the quality of satellite imagery generation. This benchmark enables standardized comparisons in the future.
    \item We conducted extensive user surveys with experts and the general audience on the representativeness of land use, constraints, and realism of images. The generated images received similar scores on all identified aspects and are favored more frequently than the real ones when the users are asked to select the more representative image.
\end{enumerate}

\section{Related Work}
\label{sec:4-lit-rev}

\subsection{Urban Imagery in Planning and Design}
The significance of visualizing urban imagery has been widely recognized in the realm of urban planning and design. Urban imagery is critical in the initial design stages and can improve public understanding and participation to facilitate effective communication and consensus building \citep{Lynch_2008, Batty_Chapman_Evans_Haklay_Kueppers_Shiode_Smith_Torrens_2000}. 
In the last decade, the scientific community began to harness the power of imagery for predictive purposes. Both satellite imagery and street view imagery were shown to be correlated with various sociodemographic and economic indicators \citep{Jean2016, Ayush2020, Rolf_Proctor_Carleton_Bolliger_Shankar_Ishihara_Recht_Hsiang_2021, Yeh2020, Gebru2017}. 
While predictive models offer significant insight into the relationship between urban imagery and the underlying factors, the recent emergence of generative models may revolutionize how we envision and build our urban environments. GenAI capitalizes on the advantages of deep learning to produce coherent natural language descriptions and vivid urban imagery. This capability offers a powerful means of enhancing communication, making complex concepts more accessible. Thus, GenAI is well-positioned to shape a future where urban development is both visionary and data-driven.

\subsection{Image Generation Models}

With the rise of deep learning and neural networks, tremendous progress has been made with image synthesis. The paradigm has shifted multiple times, from variational autoencoders (VAE) to generative adversarial networks (GAN), and now to diffusion models.
As a baseline, VAE enables sampling capabilities by imposing a Gaussian prior on the latent space \citep{ha2017neural}. However, VAEs have difficulty in generating high-quality images.
On the other hand, GAN is known for generating high-quality, realistic images. 
GAN consists of two networks, a generator and a discriminator. A game-theoretic (adversarial) training scheme updates the two networks in alternate steps, 
leading to the generation of highly realistic images \citep{larsen2016autoencoding, Berthelot2019UnderstandingRegularizer, oring2020autoencoder}. 
However, GAN's training scheme has two inherent challenges: unstable training and mode collapse  \citep{Saxena_Cao_2021}. 

In recent years, diffusion models have emerged as a more powerful paradigm in image synthesis. Diffusion models are a class of likelihood-based models that generate images by gradually removing noise from a signal \protect\citep{Ho_Jain_Abbeel_2020, Nichol_Dhariwal_2021, Dhariwal_Nichol_2021}. Compared to GANs, diffusion models exhibit better scalability and parallelization, as well as more stable training and higher fidelity images. The only drawback is that diffusion models take up more computational resources and time at both training and inference\protect\citep{Croitoru_Hondru_Ionescu_Shah_2023}. A key advancement is the latent diffusion model (LDM) \protect\citep{Rombach_Blattmann_Lorenz_Esser_Ommer_2022}, which performs denoising in a learned latent space rather than pixel space, significantly reducing computational costs while preserving image quality. Building on this, text-to-image diffusion models have been developed, where a text encoder (e.g., CLIP or T5) transforms prompts into conditioning signals that guide the denoising process. State-of-the-art models include OpenAI's DALL-E2\protect\citep{Ramesh_Dhariwal_Nichol_Chu_Chen_2022}, Google's Imagen \protect\citep{Saharia_Chan_Saxena_Li_Whang_Denton_Ghasemipour_Gontijo_Lopes_Karagol_Ayan_Salimans_et_2022}, and the open-sourced stable diffusion \protect\citep{Rombach_Blattmann_Lorenz_Esser_Ommer_2022}.

Although training diffusion models from scratch demands heavy resources, their stable scalability, and rich latent space representations have inspired researchers to fine-tune diffusion models for broader applications with more accessible computational power. A major breakthrough is ControlNet\protect\citep{Zhang_Rao_Agrawala_2023}, which has provided a versatile gateway for incorporating custom multi-modal conditions beyond the base model. By embedding additional control layers capable of processing inputs beyond standard text prompts, ControlNet allows for image generation from textual descriptions combined with visual constraints. Recent advances have expanded on ControlNet’s foundation to improve alignment between visual inputs and generated results \protect\citep{li2024controlnet++} and extending the architecture to support multi-modal conditioning and generation \protect\citep{zhang2024c3net}. Despite these innovations, the potential of diffusion models—particularly ControlNet—in urban design applications remains largely unexplored.

\subsection{Generative Urban Design}
With the rapid advances in GenAI, applications of deep generative models in urban design have been widely explored. 
There are two major approaches to urban design with generative models: designing the land use configurations and designing in the pixel space. 
Land use configurations can be formulated as a longitude-latitude-channel tensor, with the channels being different land use types. 
A pix2pix model was trained to generate land use type, floor-to-area ratio, and building cover ratio from road network sketches using GAN \citep{Park_No_Choi_Kim_2023}.
To enhance the coherence of the generated plans, LUCGAN used a spatial graph to learn the representations of surrounding contexts when generating the land-use configuration tensor \citep{Wang_Wu_Zhang_Zhou_Sun_Fu_2023}. 
This model was further enhanced with spatial hierarchy, sub-area dependency, and human instructions \citep{Wang_Fu_Liu_Chen_Wang_Lu_2023}.
In addition to GANs, reinforcement learning can also be used to learn the land use configuration tensors \citep{Zheng_Lin_Zhao_Wu_Jin_Li_2023}. 
With powerful image generation algorithms, many studies focused on visual exploration have emerged. 
For example, image-to-image generative networks are trained to predict building footprint from land cover \citep{Allen-Dumas_Wheelis_Sweet-Breu_Anantharaj_Kurte_2022}. Additionally, given the street view segmentation, researchers developed tools to generate real-time rendering of satellite images \citep{Ye_Du_Ye_2022, Espinosa_Crowley_2023} and street view images for users to interact with \citep{Noyman_Larson_2020}. Designing in the pixel space makes the design easier and more intuitive in communication while compromising some functional form details.

Comprehensive reviews of current applications can be found in \cite{Hughes_Zhu_Bednarz_2021, Wu_Stouffs_Biljecki_2022, Jiang_Ma_Webster_Chiaradia_Zhou_Zhao_Zhang_2023}. 
As this field is still in its early stages, many challenges remain. 
First, current research often falls short in addressing the diverse conditions inherent in urban planning tasks, such as site constraints and design descriptions from human experts. Additionally, GAN-based methods face limitations in generation performance due to issues like mode collapse, training instability, and poor scalability to larger architectures and datasets \citep{Croitoru_Hondru_Ionescu_Shah_2023}. 
Recently, diffusion models have emerged as a more powerful GenAI alternative, showing effectiveness in various urban tasks, such as linking auditory and visual place perceptions \citep{zhuang2024hearing}, reconstructing street views \citep{kapsalis2024urbangenai}, and enhancing satellite image resolution \citep{luo2024satdiffmoe}. However, the potential of diffusion models in urban planning remains underexplored, largely due to their high data requirements, while labeled data in the urban setting is expensive and, hence, relatively scarce.

\section{Method}
\label{sec:4-method}

\begin{figure}[ht!]
    \centering
    \includegraphics[width=\linewidth]{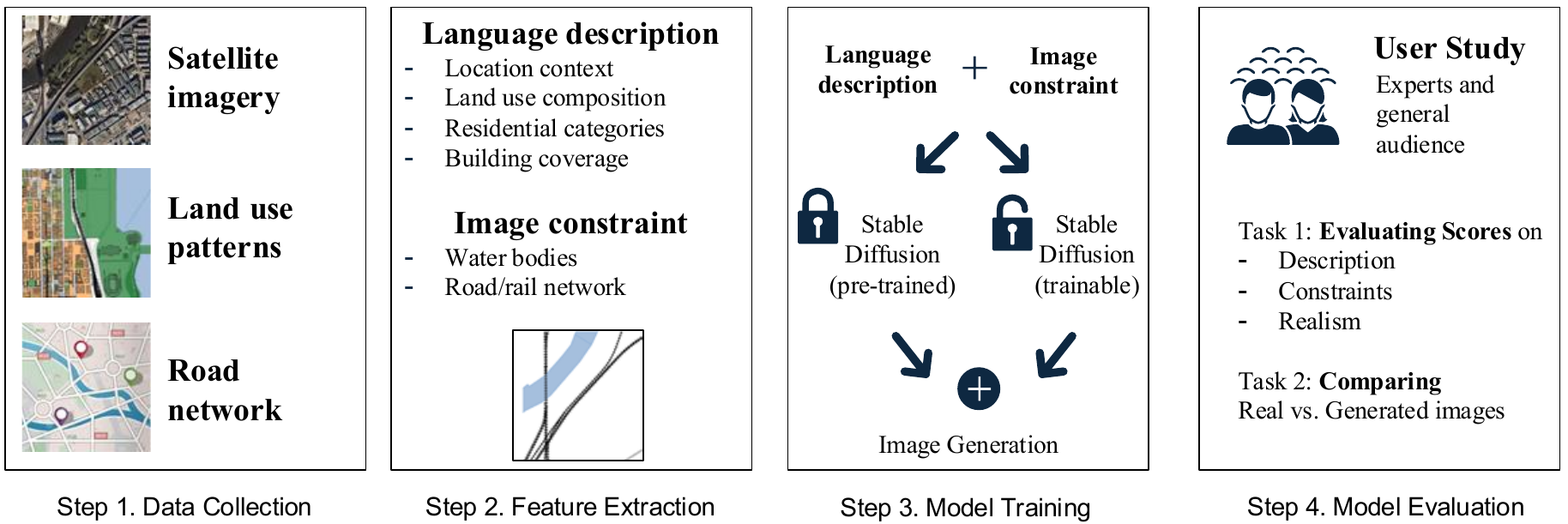}
    \caption{The proposed generative urban design workflow}
    \label{fig:4-method}
\end{figure}

This section introduces the proposed generative urban design framework, which comprises four key components: data collection, feature extraction, model training, and model evaluation (Figure \ref{fig:4-method}).
The framework begins with data collection from open-sourced, globally available satellite imagery and OpenStreetMap datasets. Spatial features are then extracted from OpenStreetMap using GIS tools, to form both environmental constraints and design descriptors. These features are spatially aligned with satellite imagery to create training pairs that serve as input for the modeling process.
In the model training phase, a stable diffusion model is fine-tuned using the ControlNet framework to generate satellite imagery informed by the extracted environmental constraints and design descriptors.
The subsequent sections provide a detailed walkthrough of each component, including methodology, tools, and implementation details.

\subsection{Data Collection}
We used publicly available datasets in this study from OpenStreetMap\footnote{\url{www.openstreetmap.org}} and Mapbox\footnote{\url{https://docs.mapbox.com/api/maps/static-tiles/}} for better generalizability and reproducibility. Satellite imagery was downloaded from Mapbox using the Slippy Map Tilenames specification \citep{OpenStreetMap_Wiki}, which defines tiles by row, column, and zoom level. To align with the 15-minute city/neighborhood concept \citep{Weng_Ding_Li_Jin_Xiao_He_Su_2019, Capasso_Da_Silva_King_Lemar_2020, Moreno_Allam_Chabaud_Gall_Pratlong_2021}, a zoom level of 16 was selected, where each tile represents a 450m x 450m area—suitable for a small mixed-use community.

\begin{figure}[h!]
    \centering
    \resizebox{0.6\linewidth}{!}{
    \includegraphics{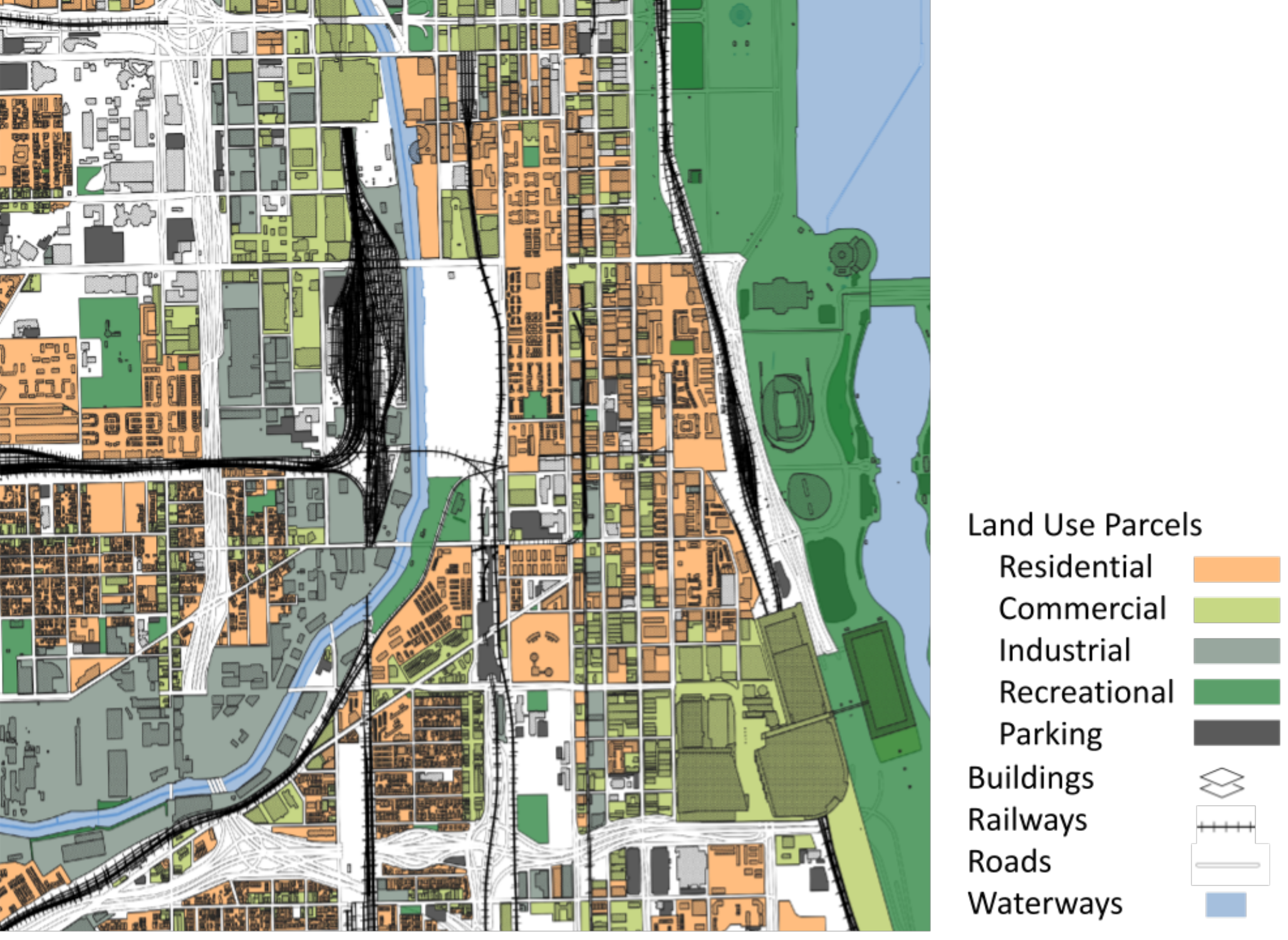}}
    \caption{Land use, transportation, and waterway layers from OpenStreetMap}
    \label{fig:4-cons-legend}
\end{figure}

We then downloaded road and land-use shapefiles from OpenStreetMap. These shapefiles include labeled land-use parcels and building footprints, categorizing areas into water bodies, residential, commercial, industrial, parks, and parking. The road layers provide information on railways and roads classified as primary, secondary, tertiary, and residential layers. Figure~\ref{fig:4-cons-legend} illustrates an overlay of these layers.

\begin{figure}[ht!]
    \centering
    \includegraphics[width=0.8\linewidth]{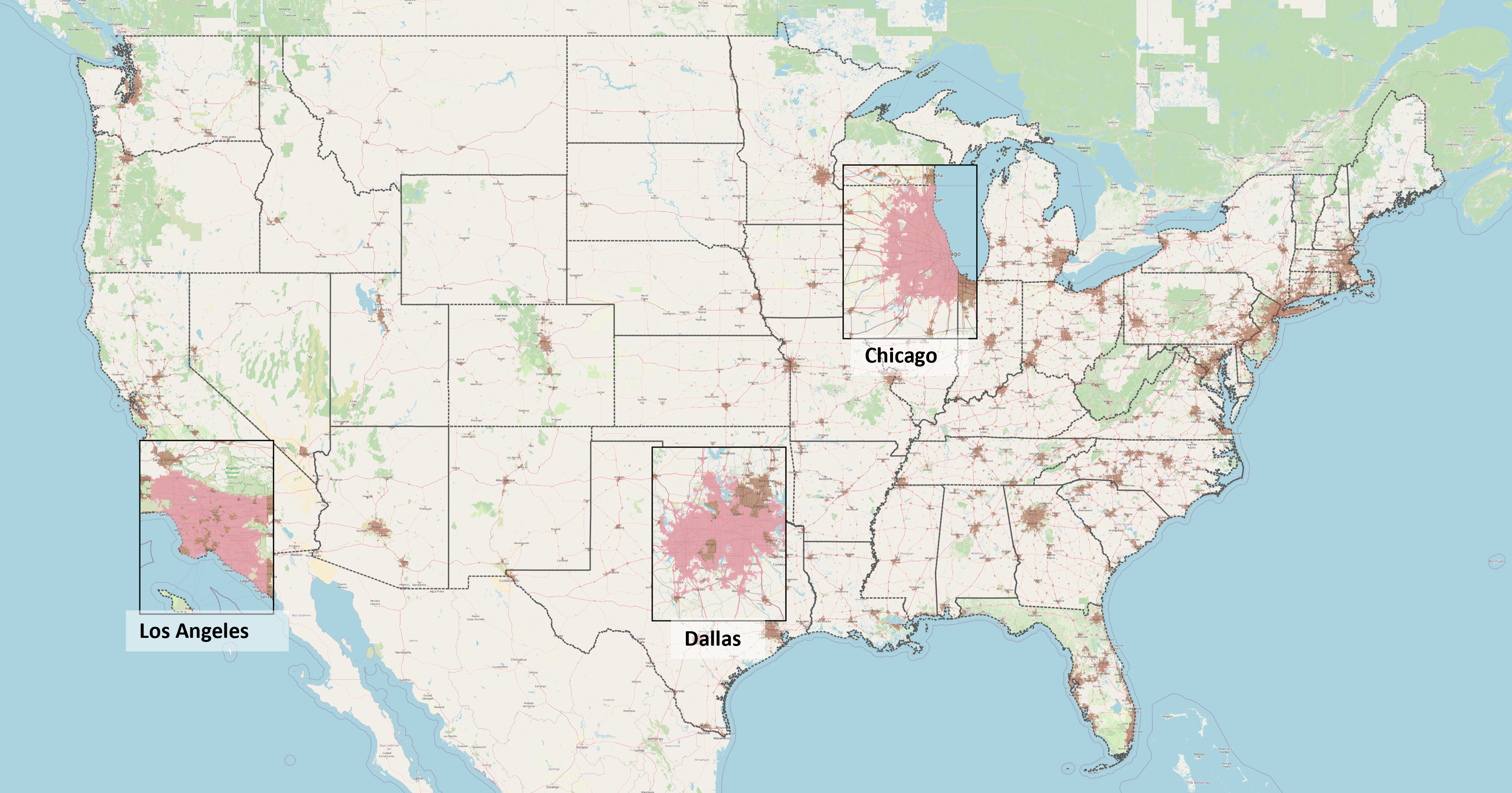}
    \vspace{-5pt}    
    \caption{The selected study areas around Los Angeles, Dallas, and Chicago}
    \label{fig:study_area}
\end{figure}

This study focuses on the urban areas defined by the U.S. Census \citep{Urban_Census} for three major U.S. cities: Chicago, Dallas, and Los Angeles (see Figure~\ref{fig:study_area}). The three metropolitan areas are similar regarding their scale, and yet differ in terms of their land use patterns. For example, Chicago metropolitan area has much more concentrated urban cores than the other two, resulting in greater variation in land use patterns across the region. The findings from the three cities can be expanded to other cities because of the global availability of OpenStreetMap and satellite imagery. 

\subsection{Data Processing}
\begin{figure}[ht!]
    \centering
    \includegraphics[width=0.8\linewidth]{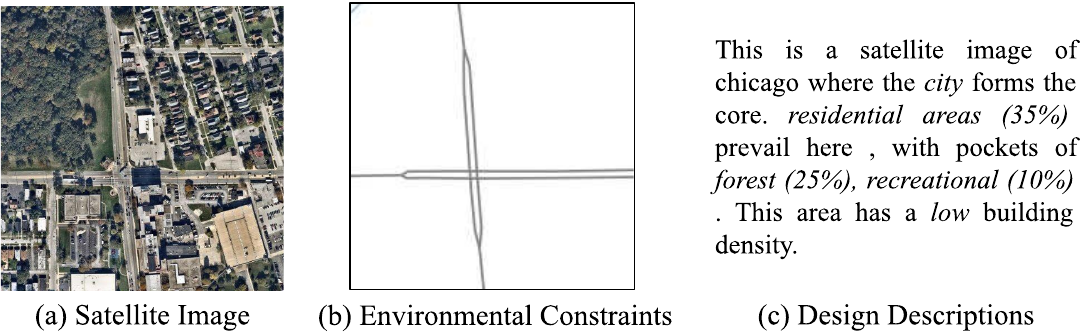}
    \caption{Illustration of a training pair}
    \label{fig:data_sample}
\end{figure}

As shown by Figure~\ref{fig:data_sample}, every satellite image tile corresponds to an environmental constraint image and a land use description by aligning the locations of multiple data sources. Environmental constraints refer to the spatial features such as road infrastructure and natural environments that remain constant during the planning process. These constraints can provide the design context while avoiding excessive restrictions. In this study, we identified railways, major roads, waterways, and land use as key imagery constraints. These layers were extracted from OpenStreetMap and processed to align spatially with the target satellite imagery. The land use controls enable us to design urban landscape while aligning with real-world planning constraints.

Design descriptions refer to the text statements that specify geographic characteristics in a structured approach. Each statement combines four components - location context, land use composition, residential type, and building coverage - through a template randomizing phrase variations. Three categories of language prompts are designed to investigate the effects of prompting style, including: (1) minimal prompt: concise prompts containing all information in bullet-point format. (2) structured prompt: descriptions are generated from templates with language variations, and (3) elaborate prompt: LLM (Deepseek-llm-7B-chat) for enriching our minimal prompt with more descriptive languages, while keeping all numerical values intact. Across the three categories, we keep the numerical information identical, varying only the richness of textual descriptions. Details of the textual components, prompt design, prompting styles can be found in Appendix~\ref{app:LLM_prompt} - Prompt Design.

One challenge in using the LLM-enriched elaborate prompts is that the text encoder used by the CLIP encoding of the stable diffusion model accepts a maximum of 77 tokens per input. Since the elaborate prompts almost always exceed this limit, we adopt a strategy commonly used by practitioners: the prompt is divided into smaller chunks, each processed independently, and the resulting embeddings are averaged to form a single final embedding \citep{moonytunes2024breaking, openai2023embedding}.

We implemented additional data preprocessing strategies to improve data completeness, augment dataset, and address sample imbalancedness. We first complemented the missing landuse classification using building classification whenever possible. In addition, to mitigate potential data incompleteness in OpenStreetMap's crowd-sourced data, we retained only tiles with over 70\% area coverage by major land-use patterns, yielding 12K, 6K, and 6K training samples for Chicago, Dallas, and Los Angeles, respectively. To improve model generalization and correctness, we applied spatial augmentation by shifting tiles along both horizontal and vertical axes. Since the original dataset contained substantially more samples from Chicago than from Dallas or Los Angeles, we used different augmentation strategies: Chicago tiles were duplicated once (2×), while Dallas and Los Angeles underwent additional shifts to achieve a 4× augmentation. This resulted in a more balanced dataset with 28K training and 2K validation samples for Chicago, 23K training and 1.7K validation samples for Dallas, and 28K training and 2.1K validation samples for Los Angeles.

\subsection{Model Training}
We fine-tune a stable diffusion model to generate satellite imagery based on environmental constraints and land use descriptions. Diffusion Models are widely used in image generation due to their stability, scalability, and adaptability, with particular advantages over the GAN model family. Inspired by physical diffusion, the stable diffusion models transform an image through a forward diffusion process, progressively adding Gaussian noise over $T$ steps to produce a noisy image $z_T$. 
The reverse diffusion process, trained using a neural network parameterized by $\theta$, learns to recover the original image $z_0$ from $z_T$ by iteratively removing noise $\epsilon_t$. Conditioning vectors $\tau_{\theta}(x)$, derived from text prompts or labels $x$, guide the generation process. The objective function for training diffusion models is:

\begin{equation}
    L_{(\theta)} = \mathbb{E}_{z_0, t, \epsilon \sim \mathcal{N}(0,1)}\left[ \| \epsilon - \epsilon_{\theta}(z_t, t, \tau_{\theta}(x)) \|^2 \right].
\end{equation}

While vanilla stable diffusion generates images based on text prompts \citep{Rombach_Blattmann_Lorenz_Esser_Ommer_2022}, it cannot capture the nuanced requirements of land use and urban design. To address this, we employ ControlNet, a framework that enhances diffusion models with custom conditions for specific applications \citep{Zhang_Rao_Agrawala_2023}. ControlNet preserves the high-quality output of the base model while enabling fine-tuning for custom new conditions. The ControlNet framework duplicates the neural network blocks that generate images: one ``locked'' copy and one ``trainable'' copy. The ``locked'' copy preserves the weights from a production-ready diffusion model, ensuring that high-quality images can be generated even from the beginning. The ``trainable'' copy gradually learns the custom condition. The final generation is a weighted combination of both copies through a ``zero-convolution" mechanism, which is a $1 \times 1$ convolutional layer with both weight and bias initialized to zeros.
\begin{figure}[ht!]
    \centering
    \includegraphics[width=0.45\linewidth]{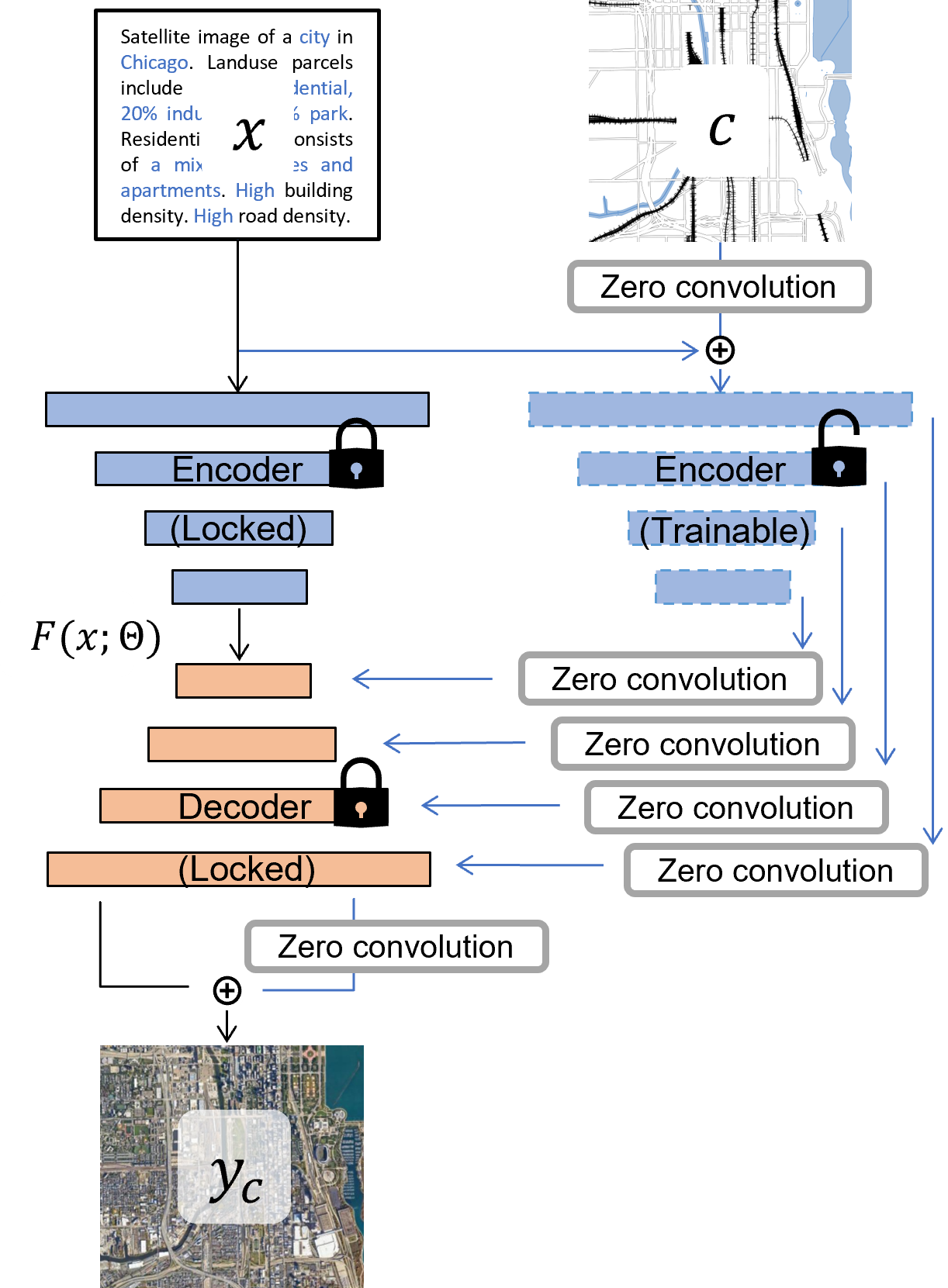}
    \caption{ControlNet Architecture}
    \label{fig:control_arch}
\end{figure}

Mathematically, the trained neural network blocks in stable diffusion are denoted as $\mathcal{F}(\cdot; \Theta)$, with trained parameters $\Theta$ that maps input text prompt $x$ to output images $y$: $y=\mathcal{F}(x; \Theta)$. During training, the network parameters $\Theta$ are fixed. For the network to learn our design descriptions and environmental constraints, we create a fresh, trainable copy of  $\mathcal{F}(x; \Theta_c=\Theta)$ and serve the combined feature vectors of our engineered design descriptions $x$ and custom environmental constraint $c$. Two instances of zero convolutions are applied by first adding on the custom condition $\mathcal{Z}(\cdot;\Theta_{z1})$ to the text prompt $x$, and second combining the custom-conditioned output of the trainable copy $\mathcal{Z}(\cdot;\Theta_{z2})$ with the output of the locked copy. Then the ControlNet output $y_c$ is

\begin{equation}
    y_c = \mathcal{F}(x; \Theta) + \mathcal{Z}(\mathcal{F}(x+\mathcal{Z}(c;\Theta_{z1});\Theta_c);\Theta_{z2}).
\end{equation}

The ControlNet is trained with the following learning objective with custom conditions $c$ at each diffusion step t:
\begin{equation}
    L_{(\Theta_c,\Theta_{z1},\Theta_{z2})} = \mathbb{E}_{z_0, t, c, \epsilon \sim \mathcal{N}(0,1)}\left[ \| \epsilon - \epsilon_{\Theta_c,\Theta_{z1},\Theta_{z2}}(z_t, t, c) \|^2 \right].
\end{equation}

At beginning, the weights of both zero convolution layers $ \Theta_{z1},\Theta_{z2}$ are initialized to 0, and the output will strictly come from the production-ready diffusion model $\mathcal{F}(x; \Theta)$: $y_c=y$. As training progresses, the trainable copy adapts to the custom conditions, enabling nuanced satellite imagery generation tailored to urban design requirements. Computationally, the model is trained for 10 epochs on a single NVIDIA V100 GPU with 32GB RAM for about 50 GPU hours.

\subsection{Quantitative and Qualitative Model Evaluation}\label{sec:4-evaluation}
To quantitatively assess the fidelity of generated satellite images, we adopt two widely used evaluation metrics: Fréchet Inception Distance (FID) \citep{heusel2017gans} and Kernel Inception Distance (KID) \citep{binkowski2018demystifying}. These metrics enable systematic and reproducible comparisons between different conditioning strategies and prompt complexities. FID measures the similarity between the distribution of real and generated images in the feature space of a pre-trained Inception network by computing the Fréchet distance between their multivariate Gaussian representations. Lower FID scores indicate closer alignment between the real and generated data distributions, reflecting higher visual fidelity. KID estimates the squared Maximum Mean Discrepancy (MMD) between real and generated image features using polynomial kernels. Compared to FID, KID is an unbiased estimator and is more robust with limited evaluation samples. We compute both FID and KID to provide more robust and comprehensive view into the generative quality of the images.

For systematic evaluation, we trained multiple ControlNet models to investigate the impacts of conditioning signals and prompting styles. All models are fine-tuned from pre-trained stable diffusion backbones, using identical training hyperparameters to allow fair comparison. We trained two ControlNets using different geographical conditioning images: ControlNet-Base and ControlNet-Landuse. ControlNet-Base uses control inputs consisting of only road and water overlays. However, real-world planning scenarios often impose additional constraints on land use, driven by factors such as property rights, zoning regulations, and community needs. To reflect this, ControlNet-Landuse extends the control image by shading a designated region with a land use category and extra language prompt explicitly describing the location and category of the shaded area. Compared to ControlNet-Base, ControlNet-Landuse provides stronger guidance on land use specification and visual appearance, enabling richer conditional generation capacity.

Qualitatively, we conducted extensive user surveys with experts and general audience on the representativeness of land use, constraints, and realism of the images. Evaluating generated urban plans presents unique challenges, mainly due to significant differences between generated and original images, and therefore having no standard benchmarks. The key question here is how real humans perceive the images, therefore we decided to conduct surveys to both qualitatively and quantitatively evaluate the generated images.

The user study has two parts: scoring and selection. Part 1 asks the user to score the presented image according to three criteria, and the second part reveals user preferences in real scenarios.  In the scoring part, users score an image between 1 and 5 on the image's consistency with the described land use, the degree to which existing infrastructure and natural environment are respected, and the realism of the images. Each user is randomly presented with either the real or the generated image, not both. In the selection part, a land use description and a constraint image are presented alongside both the real and generated images (unlabeled). The user is asked, ``Which image reveals an urban environment closer to the language description?''

The users are divided into two groups: experts and general audience. The experts are 23 graduate students and instructors from the Department of Urban Studies and Planning at Massachusetts Institute of Technology. We have set aside 20 mixed-use (having three or more land use types) neighborhoods for expert evaluation. Part 1 and Part 2 contain the same pool of images. The general audience was represented by Amazon Mechanical Turk workers from the US, with the goal of obtaining opinions from a larger population. The test set size was expanded to 50. The real and the generated images were scored by 9 people each, while 18 people completed the selection part. 
Our survey included 1,396 participants, capturing a broad cross-section of gender, age, and educational backgrounds. The participant pool has a relatively balanced gender distribution (61\% male, 36\% female). The majority of the respondents were young adults between ages 18 and 35 (61\%), with balanced representation from mid-career (36–55, 32\%) and senior (56+, 4\%) age groups. Educational backgrounds have a similar broad coverage, with 80\% holding a bachelor’s degree and beyond. The participants' gender, age, and education are summarized in Table \ref{tab:survey-demographics}.

\begin{table}[ht]
\centering
\begin{tabular}{lcc}
\toprule
\textbf{Category} & \textbf{Subgroup} & \textbf{Count (\%)} \\
\midrule
\multirow{2}{*}{Gender} 
  & Male   & 853 (61\%) \\
  & Female & 508 (36\%) \\
\midrule
\multirow{3}{*}{Age} 
  & 18--35   & 860 (61\%) \\
  & 36--55   & 450 (32\%) \\
  & 56+      & 61 (4\%) \\
\midrule
\multirow{3}{*}{Education} 
  & High school   & 182 (13\%) \\
  & Bachelor's    & 946 (68\%) \\
  & Postgraduate  & 165 (12\%) \\
\bottomrule
\end{tabular}
\caption{Demographic Summary of Survey Participants ($n = 1396$)}
\label{tab:survey-demographics}
\end{table}


\section{Results}
\noindent The results section consists of three subsections. Section \ref{sec:4-model-demo} demonstrates that the satellite images can be generated according to various language prompts, including land use compositions and city names. Such imagery generation achieves high fidelity and diversity, thus capable of re-imagining urban landscapes. Section \ref{sec:4-model-constraints} illustrates that the generated satellite images are consistent with natural environment and infrastructure constraints, thus enabling such AI-assisted design to condition on valid contextual information. Section \ref{sec:4-model-eval} presents both quantitative and qualitative evaluations of the generated images. We report FID and KID scores, and conduct two separate surveys targeting the general public and design experts, respectively. 

\subsection{Generating satellite images with language prompts}
\label{sec:4-model-demo}

\subsubsection{Generating images for land use compositions}
\begin{figure}[ht!]
    \centering
    \includegraphics[width=\linewidth]{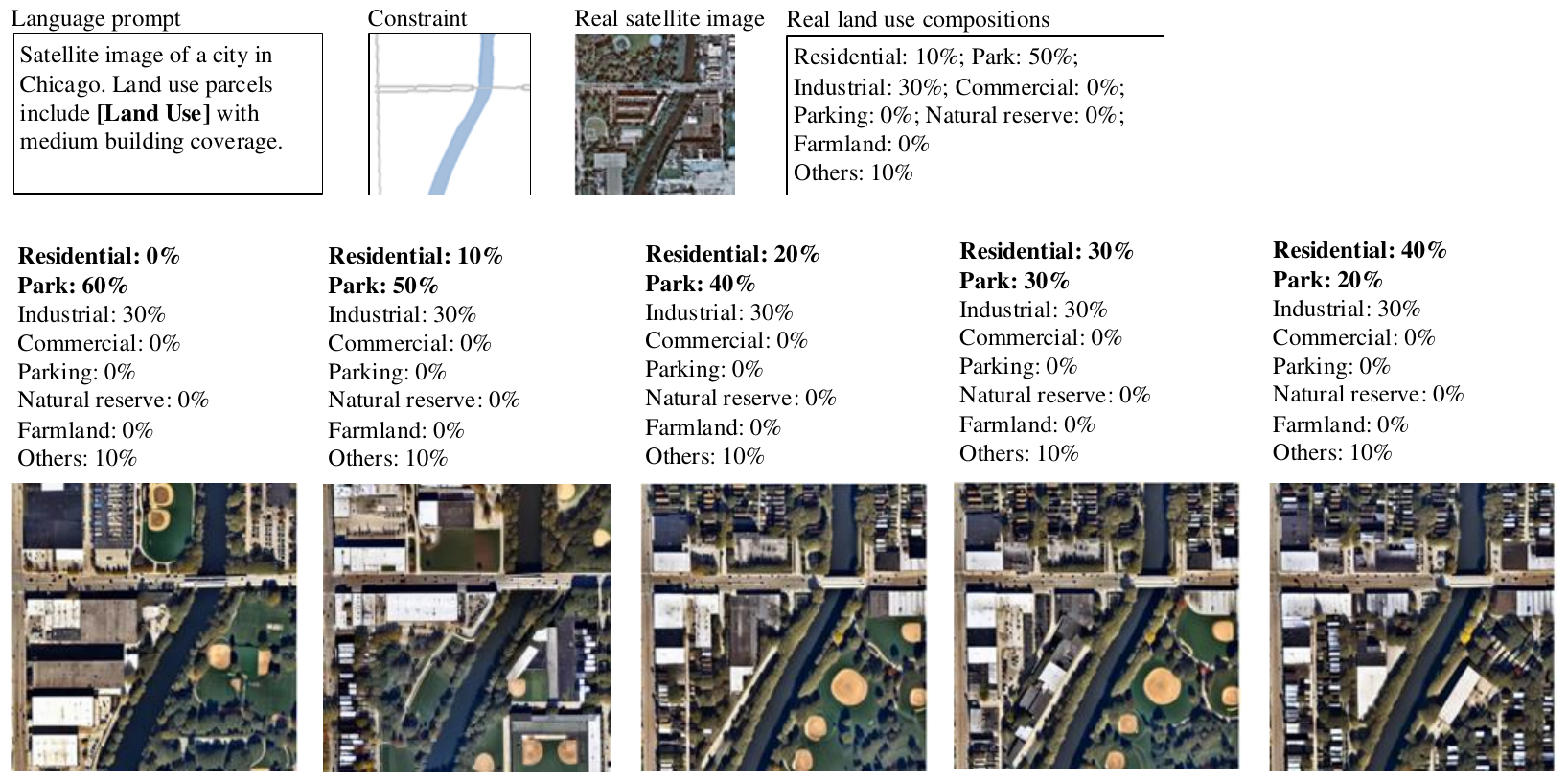}
    \caption{Generating satellite imagery with language prompts for the land trade-off between residential areas and parks in Chicago}
    \label{fig:4-gen-landuse1}
\end{figure}

We first demonstrate the model's ability to generate satellite images reflecting varying land use compositions. By controlling all other input contexts and altering only the land use composition, we observe noticeable changes in the generated images corresponding to the input land use patterns. 
Specifically, Figure~\ref{fig:4-gen-landuse1} shows an example from Chicago, where the real land use consists of 10\% residential, 50\% park, and 30\% industrial areas. In the generated images, as residential proportions increase from 0\% to 40\% and park proportions decrease from 60\% to 20\%, noticeable changes emerge. Park areas, depicted as expansive green spaces with human-built structures, visibly shrink from left to right. Concurrently, the increase in residential land use is reflected in the emergence of dense, small-block neighborhoods characterized by row houses, which become more prominent as residential proportions rise. The industrial land use remains consistent, with large structures maintaining fixed proportions. These results demonstrate the system's ability to reliably translate input land use compositions into visual representations.

In addition to reflecting land use compositions, the model has captured some spatial relationships between different land use types, even though these relationships were not explicitly defined in the prompts. In the first image, park areas are separated from industrial zones by major roads and waterways, while in the last image, parks and residential areas are closely integrated, suggesting an interaction between green spaces and housing. This suggests the model may have inferred and applied some spatial planning tendencies, such as park placement, from the training data.

\begin{figure}[ht!]
    \centering
    \includegraphics[width=.75\linewidth]{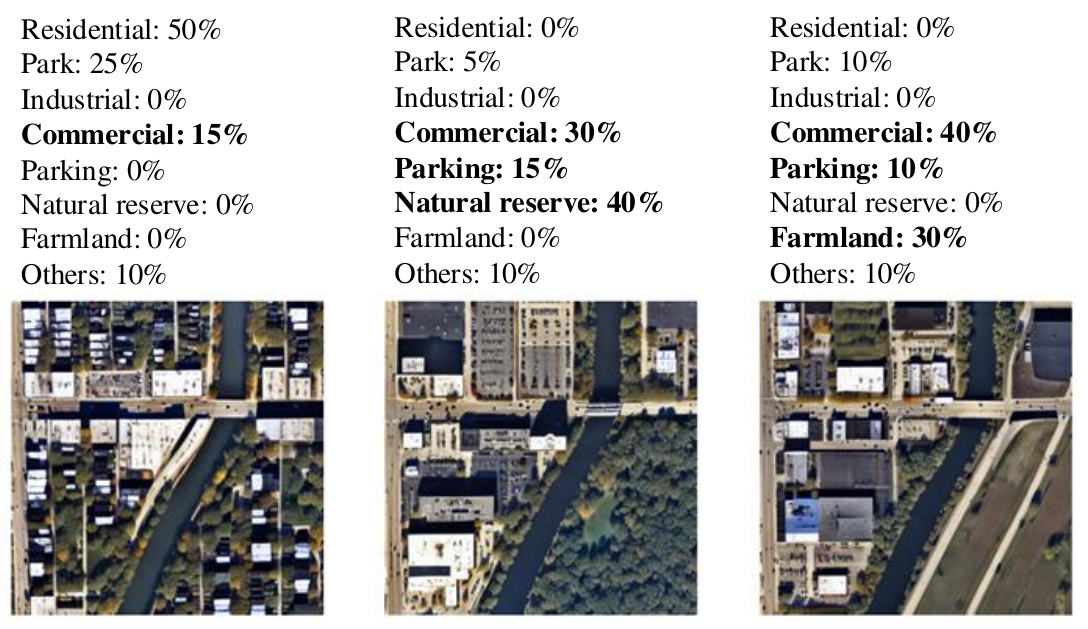}
    \caption{Generating satellite imagery with language prompts for non-existing land use labels in the original satellite imagery}
    \label{fig:4-gen-landuse2}
\end{figure}

The previous example demonstrates the model's ability to control the proportions of existing land use types in a real urban scenario. In Figure~\ref{fig:4-gen-landuse2}, we further highlight the model's capability to generate novel, unseen land use patterns within the context of an existing urban environment. This capability enables the re-imagination of diverse land use planning possibilities for a given site. The first image envisions a mixed-use neighborhood with 50\% residential and 15\% commercial land use. Large commercial blocks are aligned along the central street, while residential areas are placed along quieter side streets, creating a livelier, more accessible environment compared to the existing mixed residential-industrial layout in Figure~\ref{fig:4-gen-landuse1}. The second image features 30\% commercial buildings, 15\% open parking spaces, and 40\% natural reserves. The natural reserve is portrayed as a densely forested region with minimal human-made infrastructure. Commercial buildings are situated on the opposite side of the river, encircled by open parking spaces. This arrangement aligns with common planning principles, where parking areas support the functionality of adjacent commercial blocks, while natural reserves are typically separated from other land use types to preserve their ecological integrity. The third image depicts an area composed of 40\% commercial land, 10\% parking, and 30\% farmland. The generated image portrays farmland as vast, open brown fields devoid of visible structures. Unlike the first image, where commercial buildings are concentrated along the central main road, the commercial areas in this scenario are distributed along the left main road, with building density decreasing near the farmland. This reflects common planning principles of low-density development near agricultural zones.

In summary, the examples demonstrate the model's ability to effectively represent various land use compositions and their distinct characteristics. Beyond reliably representing these distinct land use types, the model demonstrates an ability to capture spatial relationships between them in a plausible manner, reflecting common planning tendencies observed in the training data.

\subsubsection{Generating images for learning across cities}
\begin{figure}[th!]
    \centering
    \includegraphics[width=\linewidth]{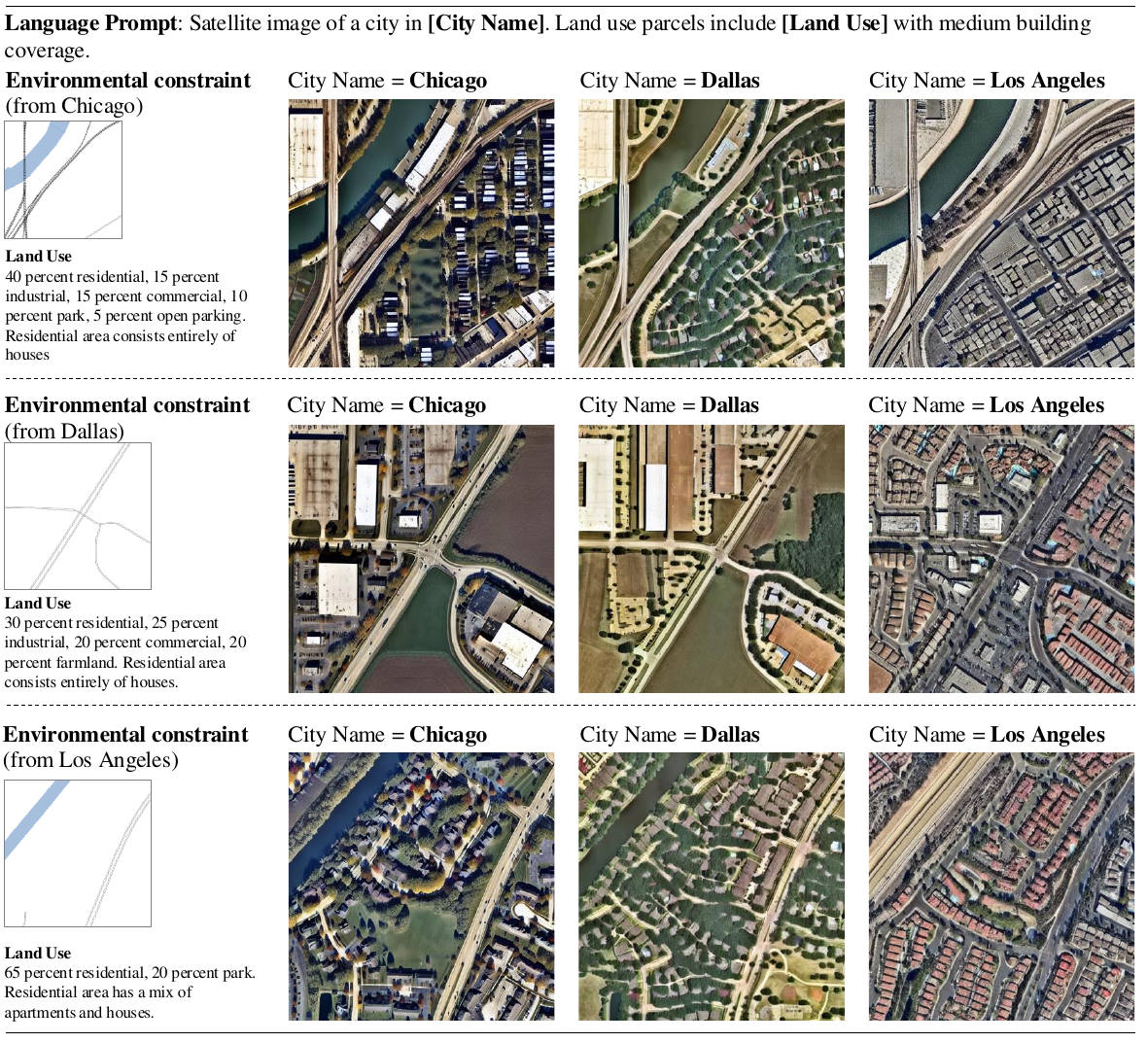}
    \caption{Generating satellite imagery across three cities conditioning on the same environmental constraint}
    \vspace{-5pt}        
    \label{fig:4-gen-cities}
\end{figure}

Urban planners often analyze urban contexts to understand the unique characteristics of different cities. Figure \ref{fig:4-gen-cities} demonstrates how our Stable Diffusion model facilitates cross-city learning by reflecting the distinct urban forms of Chicago, Dallas, and Los Angeles. The figure presents generation results for the three cities under identical constraints and land use descriptions, with only the city name varying. Despite the same inputs, the generated images display notable differences.

In the first scenario (row), the land use composition is set to 40\% residential, 15\% industrial, 15\% commercial, and 10\% park. Despite the identical constraints, the road network layouts differ significantly between cities. Chicago exhibits a strong grid-based alignment along north-south and east-west axes, reflecting its historic planning tradition. In contrast, Dallas and Los Angeles display less rigid layouts, with more diagonal and curvilinear streets. Additionally, green spaces and tree coverage are more prominent around residential areas in Chicago and Dallas. In Los Angeles, however, the built environment is characterized by denser building arrangements with narrower spacing between structures. Buildings are located closer to the streets, resulting in a compact and car-oriented urban form with limited green buffers. 

In the second scenario (row), the land use composition is 30\% residential, 25\% industrial, 20\% commercial, and 20\% farmland. In Chicago and Dallas, the generated road networks are relatively sparse, with large vacant areas representing farmland and sizeable building blocks surrounded by open parking lots, reflecting industrial and commercial zones. In contrast, Los Angeles exhibits a denser road network, with mid-sized building blocks lining the main streets for industrial and commercial areas. Residential zones in Los Angeles are depicted as small, tightly packed building blocks, closely integrated with the surrounding road networks. Additionally, these residential areas often feature swimming pools, visible as light blue spots, highlighting the city's warmer climate and cultural inclination toward private leisure spaces.

In the third scenario (row), a residential neighborhood is imagined, with 65\% residential and 20\% park. In Chicago, the park is depicted as a centralized, expansive green space embedded within the residential neighborhood. In Dallas, green spaces are more scattered, surrounding individual houses and apartments, while in Los Angeles, they are distributed along the streets. Additionally, the roads in Los Angeles are visibly wider than in the other two cities, further emphasizing the city's car-centric urban form. The treatment of the riverbank also varies across cities: in Chicago and Dallas, the river is flanked by green spaces and trees, while in Los Angeles, the river is bordered by human infrastructure, reflecting a more urbanized environment. These results illustrate the model's capability to capture and replicate the unique urban styles of different cities, offering a valuable tool for envisioning alternative planning scenarios and drawing inspiration from diverse urban practices. 

\subsubsection{Generating distinct images with fixed prompts}
\begin{figure}[th!]
    \centering
    \includegraphics[width=\linewidth]{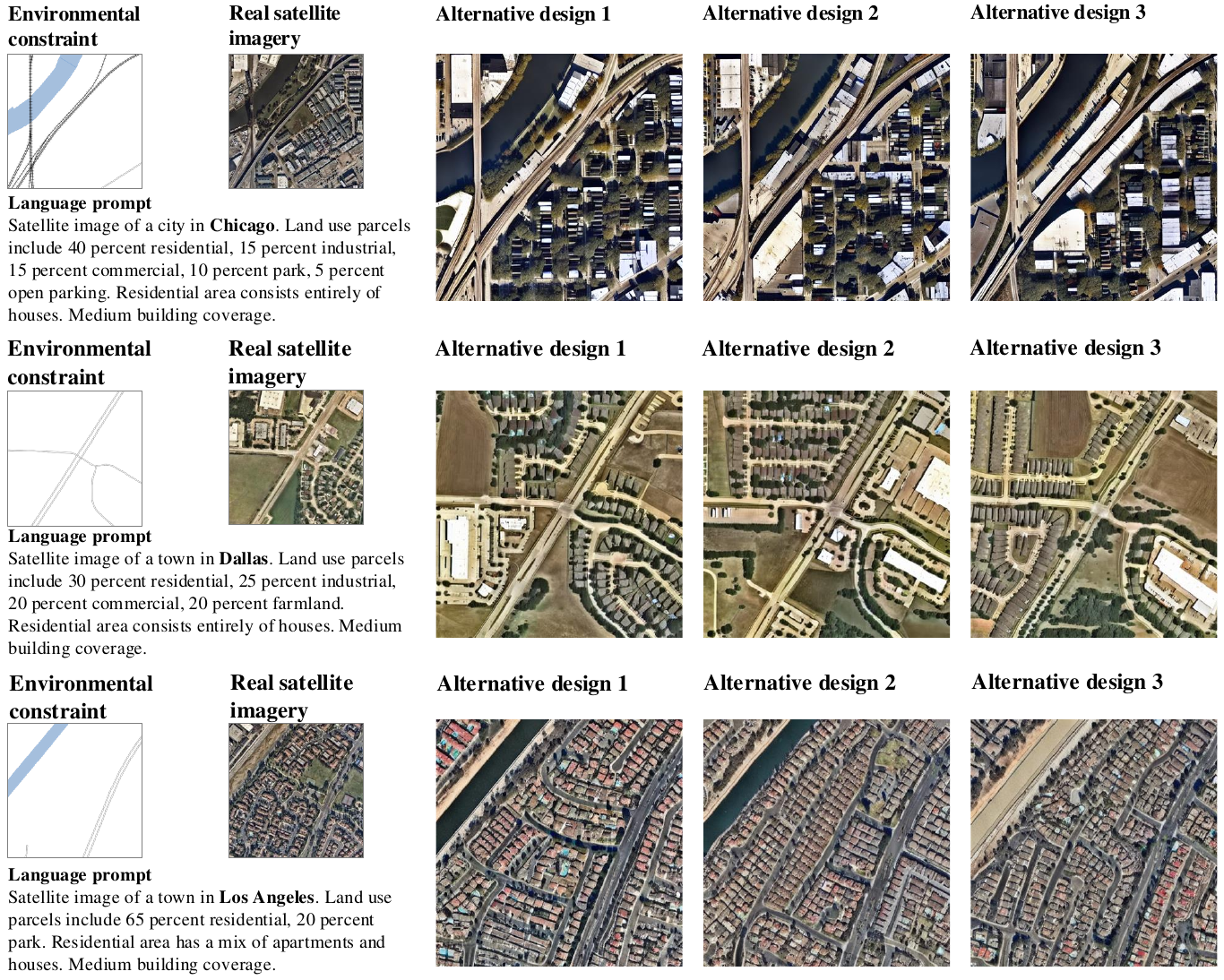}
    \caption{Generating diverse satellite imagery conditioning on fixed language and imagery prompts}
    \vspace{-10pt}        
    \label{fig:4-gen-diversity}
\end{figure}

There is a balance between precision and creativity when using AI for visualizations: specifying building functions versus allowing AI to generate with full creative freedom. Considering the model's role in inspiring the concept planning phase of real-world projects, we control land use types but let the model decide their spatial arrangement. The last two columns in Figure \ref{fig:4-gen-diversity} showcase alternative designs from the same prompt, demonstrating the model's ability to produce diverse layouts. While maintaining consistent land use components and proportions, the designs vary in spatial layout, shape, and treatment of urban elements.

In the first scenario (Chicago), the alternative designs present varied approaches to public space along the riverbank. The first design emphasizes expansive green spaces with minimal infrastructure, creating a park-like setting ideal for outdoor activities. The second design prioritizes dense buildings along the river, maximizing opportunities for commercial development but reducing open space. The third design strikes a balance, featuring a long square along the riverbank for public access while maintaining narrower infrastructure behind it to support riverside businesses.

In the second scenario (Dallas), the designs highlight different residential building layouts. The first alternative separates the residential area into two clusters, one in the northwest and another in the southeast, leaving significant vacant space nearby. The second design creates a more compact residential zone with rows of buildings concentrated in the northwest. The third design arranges residential blocks along the streets, with green spaces interspersed in the center. These variations provide multiple options for residential neighborhood planning, adaptable to specific community needs and living requirements.

In the third scenario (Los Angeles), the designs showcase diverse approaches to road network planning. The first design emphasizes multiple connections branching from the main road toward the river, enhancing accessibility to the waterfront and encouraging interaction with the riverbank. The second design prioritizes a parallel alignment of roads with the main road, with building blocks oriented along the river. The third design introduces a network of curved roads, resulting in a more organic neighborhood layout. These examples demonstrate the model's ability to generate diverse design layouts for public spaces, residential blocks, and road networks under identical input conditions, providing multiple alternatives to inspire human designers and support creative exploration in the early stages of urban planning.

These results demonstrate the model’s initial potential to support design innovation. Design innovation can be considered across three layers: (1) innovation relative to the original satellite image, (2) innovation relative to the city’s prevailing design style, and (3) the capacity to generate entirely novel, creative solutions. First, our results demonstrate that the model can generate urban layouts that differ significantly from existing real-world cases, offering new possibilities for reimagining urban environments. As illustrated in Figure~\ref{fig:4-gen-landuse2}, the framework is able to produce land use patterns that do not exist in the original satellite imagery. In Figure~\ref{fig:4-gen-diversity}, even under identical site constraints and language prompts, the model generates diverse urban design diagrams that deviate from their real-world counterparts. These findings suggest that our framework is capable of producing innovative solutions beyond simply replicating input images. Second, from a city-style perspective, we acknowledge that different cities exhibit distinct planning conventions, and our framework tends to capture and reflect these stylistic norms. We also demonstrate the potential for cross-city style transfer, which allows one city to be reimagined using the planning style of another. As shown in Figure~\ref{fig:4-gen-cities}, our model successfully generates satellite imagery for three different cities based on identical input constraints, resulting in distinct urban designs that diverge from each city’s original stylistic patterns. This provides a promising pathway for encouraging innovation beyond the bounds of existing urban design norms. Finally, regarding the generation of entirely original and creative solutions, we believe this remains an open question. Evaluating the creativity of GenAI-generated designs against that of human designers is a valuable direction for future research. This would involve developing new evaluation metrics and possibly human-in-the-loop systems to foster truly imaginative and context-sensitive urban solutions.

\subsubsection{Generating images with three prompting styles}
\begin{figure}
    \centering
    \includegraphics[width=\linewidth]{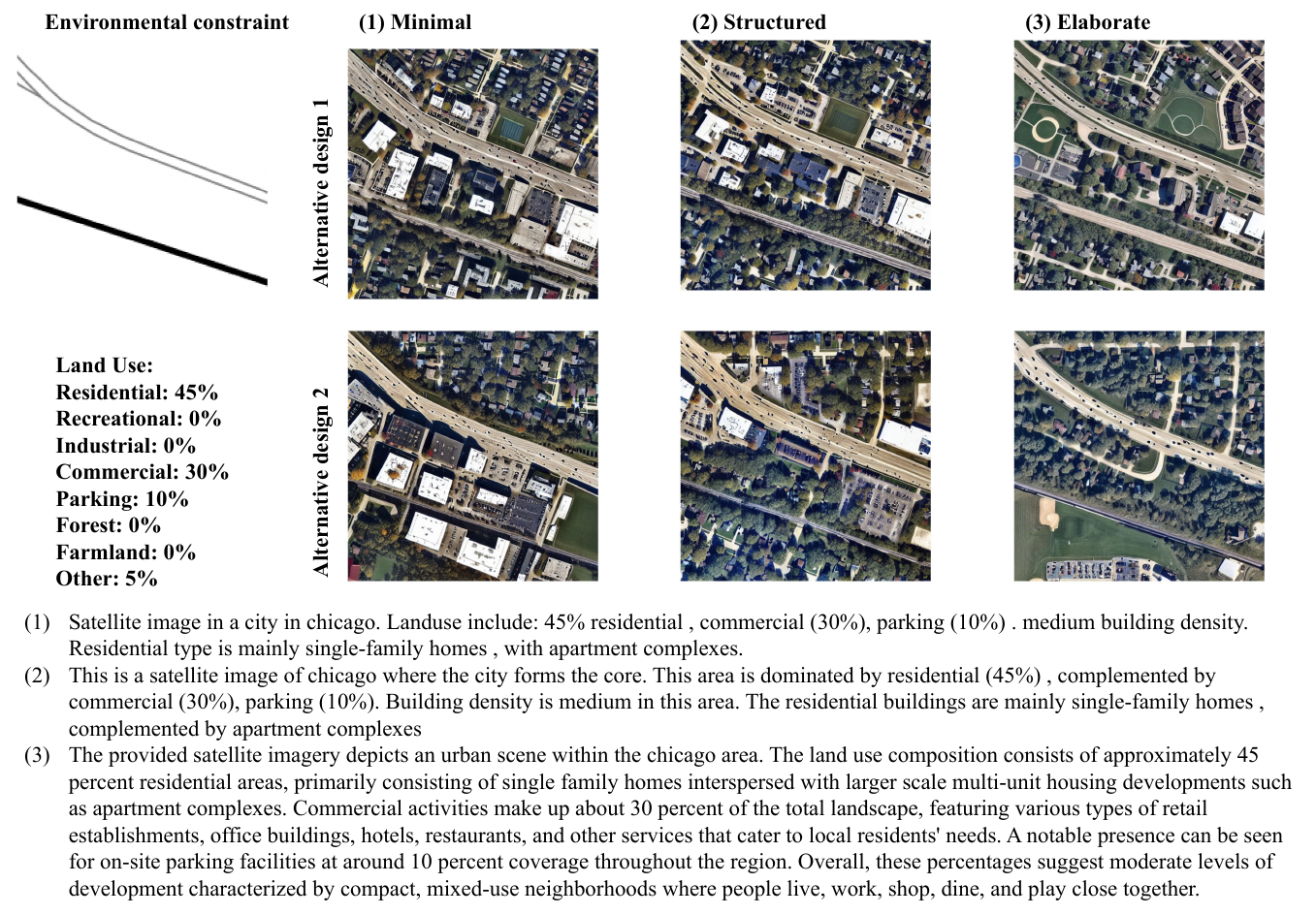}
    \vspace{-10mm}
    \caption{Generating satellite imagery with three prompting styles}
    \label{fig:4-gen-language}
\end{figure}

To ensure robustness and flexibility in real-world applications, we trained models using three prompting styles. Figure~\ref{fig:4-gen-language} illustrates generation results conditioned on the same information but different language formats. For each case, two alternative outputs are shown with identical constraint images and land use descriptions. Overall, models trained with minimal and structural prompts achieve more accurate representation of the specified land use mixes. In contrast, models trained with elaborate prompts sometimes generate land use compositions that deviate from the specified parameters. This discrepancy may arise because the elaborate prompts, enriched by the LLM, introduce additional general descriptions that are not directly tied to the specific tile. As a result, the overall informational precision of the prompt is reduced, weakening ControlNet’s ability to accurately generate land use patterns. Between the minimal and structural prompts, there is little visual difference in the generated images, suggesting that our approach can effectively interpret both bullet-point formats and natural language descriptions.

\subsection{Generating images under constraints}
\label{sec:4-model-constraints}
\begin{figure}
    \centering
    \includegraphics[width=\linewidth]{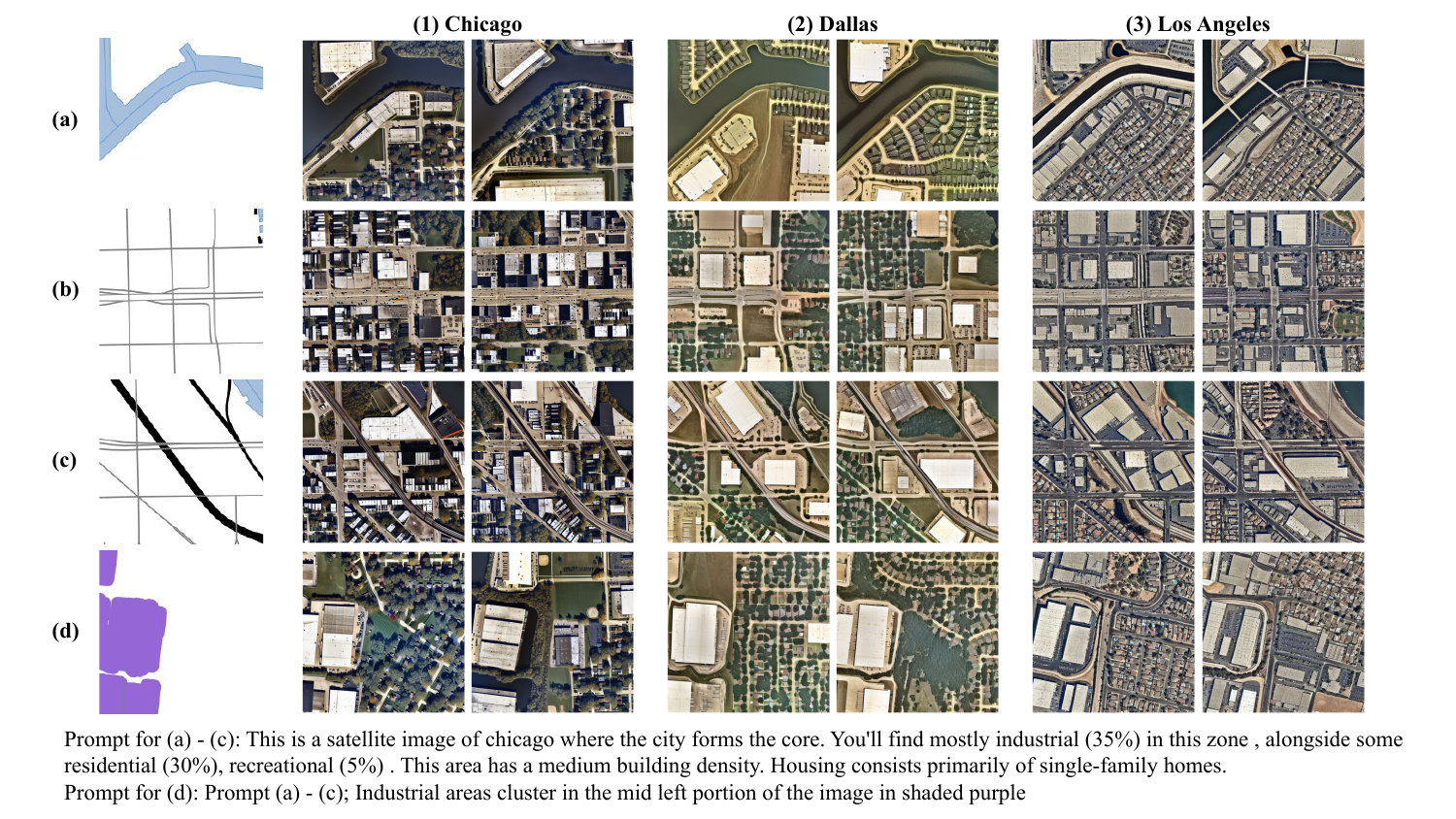}
    \vspace{-5mm}
    \caption{Generating satellite imagery with various environmental constraints: (a) waterway (b) roads (c) railway (d) industrial land use }
    \label{fig:4-gen-constraints}
\end{figure}

The stable diffusion model can not only generate urban landscapes according to text descriptions as in Section \ref{sec:4-model-demo}, it can also generate landscapes according to various imagery inputs. As shown in Figure~\ref{fig:4-gen-constraints}, the model outputs consistently align with the environmental constraints as waterways, road networks, railways, and land use parcels. 

The first example (row) highlights the model’s responsiveness to waterways across the three cities in our experiments. The generated designs effectively integrate riverbank features, such as green spaces along the river and bridges going across, highlighting tailored urban responses to the waterway. The second example (row) illustrates the influence of road constraints on the generated satellite images. Across all three cities, the urban plans align with the road network while producing distinct building layouts. 
In Chicago, the output features a dense urban form, with tall, large buildings arranged along the gridded street network. In Dallas, the model produces a more rural visual character, where industrial blocks, residential buildings, and parking lots are interspersed with green spaces. In Los Angeles, the result showcases a mixed-use urban form, combining small residential buildings, park spaces, and medium-sized high-rise structures. The third example (row) focuses on railway constraints and their impact on the generated designs. Beyond accurately capturing the location and shape of the railways, the model adjusts the surrounding urban forms in response to these constraints. Buildings near the railway are either shaped to align with its orientation, or place large vacant spaces adjacent to the railway, creating a buffer zone. The fourth example (row) highlights the use of additional land use controls, reflecting scenarios where specific development requirements must be preserved, such as protecting existing buildings from demolition. In this case, an industrial land use—representing utility or service buildings—is specified within a designated area. Across all generated samples, the model consistently retains industrial buildings within the controlled region while planning the surrounding urban fabric accordingly. This demonstrates the model’s ability to reflect localized land use constraints, ensuring that protected zones are preserved and new developments are sensitively integrated. Such capabilities are essential for real-world applications where development must accommodate legacy infrastructure or adhere to strict zoning requirements. 

\subsection{Evaluation}\label{sec:4-model-eval}

\subsubsection{Quantitative evaluation}

Table~\ref{tab:fid_score} and Table~\ref{tab:kid_score} report FID and KID scores for the 5700 validation images across three cities, comparing ControlNet-Base and ControlNet-Landuse under varying prompt complexities (Minimal, Structural, and Elaborate). These results provide several key insights.

\begin{table}[th!]
    \centering
    \begin{tabular}{c|c|c|c|c}
        \toprule
        \multirow{2}{*}{City} & \multicolumn{3}{c|}{ControlNet-Base} & \multirow{2}{*}{ControlNet-Landuse} \\
        & Minimal & Structural & Elaborate &  \\
        \midrule
        Overall & 68.08 & 63.15 & 66.19 & 58.94 \\
        \midrule
        Chicago & 95.73 & 96.96 & 102.05 & 91.44 \\
        Dallas & 113.30 & 94.78 & 108.13 & 84.74 \\
        Los Angeles & 70.83 & 76.22 & 64.55 & 77.38\\
        \bottomrule
    \end{tabular}
    \caption{Fidelity performance (FID score) comparison of generated images}
    \label{tab:fid_score}
\end{table}

\begin{table}[h!]
    \centering
    \begin{tabular}{c|c|c|c|c}
        \toprule
        \multirow{2}{*}{City} & \multicolumn{3}{c|}{ControlNet-Base} & \multirow{2}{*}{ControlNet-Landuse} \\
        & Minimal & Structural & Elaborate &  \\
        \midrule
        Overall & 0.04467 & 0.03857 & 0.04488 & 0.03514 \\
        \midrule
        Chicago & 0.06874 & 0.06990 & 0.07022 & 0.06068\\
        Dallas & 0.09893 & 0.07702 & 0.08591 & 0.06366\\
        Los Angeles & 0.05012 & 0.05914 & 0.04143 & 0.06057\\
        \bottomrule
    \end{tabular}
    \caption{Fidelity performance (KID score) comparison of generated images}
    \label{tab:kid_score}
\end{table}

Overall, ControlNet-Landuse consistently outperforms ControlNet-Base, achieving the lowest overall FID (58.94) and KID (0.03514). This confirms the benefit of incorporating detailed and accurate semantic information, a single shaded land use region, into the training process. By comparison, ControlNet-Base, which conditions only on roads and water, and a description of the proportions of the landuse types, has to learn the appearance of and distinguish multiple landuses in one image, hence yielding higher FID and KID scores.
Within ControlNet-Base, the structural prompt format yields the best fidelity scores (FID 63.15 and KID 0.03857). This result suggests that providing moderate contextual detail strikes a balance between under-specification (Minimal) and potential over-specification and complex language with the same underlying information (Elaborate). 

We observe that per-city FID and KID scores are generally higher (worse) than the overall averages. This is expected and can be attributed to dataset aggregation effects. The overall metrics are computed by pooling all generated samples across cities, which increases sample diversity and may smooth over localized artifacts or outlier distributions. In contrast, city-specific evaluations isolate smaller subsets with more uniform visual and structural characteristics, which can accentuate generation errors and reduce diversity in feature space and both factors negatively affect FID/KID. This discrepancy between the average and per-city performance also reflects intra-city complexity: within a single city, urban patterns often include tightly clustered styles (e.g., dense grid layouts, homogeneous suburbs), where generation errors become more statistically distinguishable. Meanwhile, when cities are evaluated together, cross-city variance dilutes the impact of any individual anomaly, resulting in lower aggregate scores. These findings reinforce the importance of per-city breakdowns in benchmarking, as they reveal performance gaps that may otherwise be masked in global averages and help identify where models are more or less robust to distinct urban contexts.

In our experiments, fidelity performance varies substantially across cities. Dallas consistently produces the highest FID and KID values across all configurations, suggesting greater difficulty in generating plausible imagery for this region. This may be due to lower visual consistency within the Dallas training data or more complex, fragmented urban morphology. In contrast, Chicago and Los Angeles exhibit generally lower and more stable scores. However, Los Angeles frequently demonstrates opposite trends compared to the other cities in model feature comparisons: ControlNet-Base with more elaborate prompts achieves better fidelity than ControlNet-Landuse. This result highlights that city-specific features, such as heterogeneous land use patterns and less rigid urban structure, can interact differently with model design choices. As such, establishing per-city evaluation metrics is critical to capturing these nuanced behaviors, ensuring that model improvements are not evaluated solely on aggregate performance but are tested for robustness across varying urban environments.

\subsubsection{Qualitative evaluation through user study}
As discussed in Section~\ref{sec:4-evaluation}, a user study was conducted to gather evaluations and feedback on the generated urban imagery compared to real one. The study consisted of two parts: scoring and selection. In Part 1, participants were asked to score the images based on their alignment with site constraints, design descriptions, and overall realism. In Part 2, participants were asked to choose between real and generated images, selecting which one better matched the urban environment and design description.

Results show that the generated images have successfully learned the features from real images. Table~\ref{tab:4-eval_metrics} presents the results for Part 1 of the user study, tabulating the scores in the format of ``expert $|$ general''. In general, the experts can better tell the difference between generated and real images. The experts gave generated images slightly lower scores on matching land use patterns (-0.34 points) and realism (-0.81 points) while identifying them as better conforming to the constraints (+0.26 points). The scores given by the general audience between real and generated are almost the same. On average, the generated images received scores 0.1 lower on land use and 0.04 lower on constraints while they appeared equally real compared to the real images. 

\begin{table}[ht!]
    \centering
    \begin{tabular}{cccc}
        \toprule
         Expert $|$ General & Design Description & Site Constraint  & Realism \\
         \midrule
         Real&  3.73 $|$ 3.77 &  3.68 $|$ 4.17 &3.87 $|$ 3.78\\
         Generated & 3.39 $|$ 3.67 & 3.94 $|$ 4.13 & 3.06 $|$ 3.78 \\
         \midrule
         Difference&  -0.34 $|$ -0.10 &  +0.26 $|$ -0.04 & -0.81 $|$ 0.00\\
         \bottomrule
    \end{tabular}
    \caption{User study scores of generated and real satellite images (Min score:1, Max score:5)}
    \label{tab:4-eval_metrics}
\end{table}

\begin{figure}
    \centering
    \includegraphics[width=0.6\linewidth]{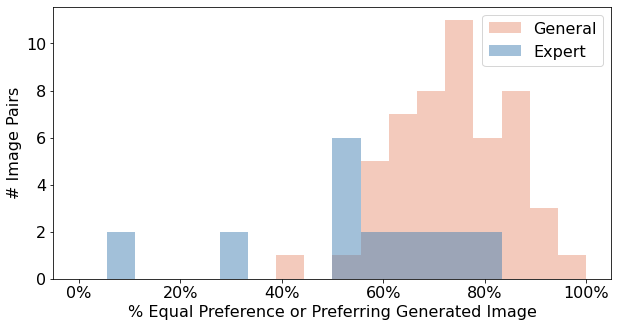}
    \caption{Distribution of votes for each pair of images}
    \label{fig:4-select_hist}
\end{figure}

Figure~\ref{fig:4-select_hist} presents the results from Part 2 of the user study, illustrating the distribution of votes for each pair of images. The findings reveal that both user groups—experts and the general audience—favored the generated images over real images, as the generated images better aligned with the provided descriptions. Based on majority votes, 12 out of 20 (60\%) and 42 out of 50 (84\%) generated satellite images were preferred over real images by the expert and general groups, respectively. Our analysis highlights a notable divide in preferences between the two groups. Experts displayed a more balanced split, with a median of 56\% favoring the generated images in each pair. In contrast, the general audience demonstrated a significantly stronger preference for the generated images, with a median of 75\% favoring them. This divergence in preferences underscores differing evaluation criteria: experts may focus on technical accuracy and conceptual fidelity, while the general audience appears more influenced by visual appeal and descriptive alignment.


\section{Conclusion and Discussion}
\label{sec:4-conclusion}
The iterative nature of urban planning favors tools that streamline the planning process, enhance communication, and provide quick feedback. AI has recently shown promising urban planning capabilities, such as generating data-driven insights, assessing and improving plan performance, and crafting visualizations. Despite AI's significant potential to revolutionize urban planning, practical implementation faces numerous challenges. Current research is primarily geared towards narrowly defined, tightly controlled tasks. Urban design, with its variety and complexity, does not lend itself easily to simple parameterization. Additionally, generative AI's reliance on substantial data volumes poses a challenge, as acquiring labeled data in urban contexts is costly and thus limited. 

This study tackles these challenges by introducing a GenAI framework for urban planning, leveraging ControlNet and stable diffusion. The framework is realized through a model trained on data automatically labeled from widely accessible resources, providing a novel approach to integrating AI into urban planning and bridging the gap between theoretical potential and practical application. The stable diffusion model generates satellite imagery based on environmental constraints and textual descriptions, allowing human-guided control over AI-generated land use patterns. This model also enables the creation of diverse urban landscapes under identical constraints and descriptions, fostering creativity in urban planning. It consistently adheres to various constraints while skillfully incorporating local textures from different cities into its designs. 


Our diffusion model offers several potential applications, including rapid visualization of conceptual designs, uncovering implicit associations and styles, and fostering public engagement in urban planning. First, the model enables near real-time visualizations of conceptual designs, allowing planners and the public to explore ``what-if'' scenarios for the neighborhoods. For example, users might ask, ``What if we remove this major highway ramp from our neighborhood?'', ``What if we convert this residential area into a commercial zone?'', or ``What if we create a park in this space?'' While the AI tool does not provide detailed architectural plans, it offers a bird's-eye perspective that supports intuitive understanding and informed discussions. Second, the diffusion model can reveal implicit associations and stylistic patterns that are difficult to articulate. By learning styles from different cities, the model facilitates cross-city comparisons, offering inspiration for elements that may not have been initially included in the constraints or prompts. This ability to highlight stylistic diversity can guide planners in exploring new possibilities. Third, the tool empowers non-professionals to engage with professional urban planning concepts. By consolidating complex planning concepts into generative visuals, it enables the public to envision urban planning ideas and participate creatively in local planning initiatives. 

Despite its advancements, this study has limitations that point to two key directions for future research. First, expanding the range of inputs and outputs could significantly facilitate more nuanced representation of urban planning. Inputs such as detailed zoning information and sociodemographic data could provide a richer context, while outputs like building footprints, heights, and street views could offer more comprehensive and actionable design visualizations. Second, since the status quo does not always reflect the ideal design, it is crucial to incorporate value judgments into the generative process, such as equity, sustainability, or resilience. This would enable the creation of planning that go beyond replicating existing urban landscapes to envisioning more desirable futures. We believe the transformative potential of GenAI in urban planning will have a lasting impact, so we leave such critical avenues for future exploration. Third, while our work demonstrates the potential of GenAI in producing visually plausible satellite images, there remain notable limitations in the granularity and functionality of the generated outputs. Key urban features—such as street furniture, park layouts, and pedestrian pathways—often lack sufficient precision in their placement and form, limiting their immediate utility for real-world design applications. Furthermore, the functional quality of the generated designs has yet to be rigorously evaluated in terms of service and green space accessibility, transportation efficiency, social inclusivity, and other critical urban performance metrics. These shortcomings may be addressed by further enhancing the ControlNet framework to better capture high-resolution urban features and to deepen its understanding of the relationship between the built environment and its associated functional qualities. Fourth, we acknowledge the limitation that our current GenAI framework processes each image tile independently, with limited consideration of the multi-scale nature of urban planning and the spatial continuity between adjacent land parcels. To address this challenge, future work should explore hierarchical generation frameworks that capture multi-scale perspectives aligned with urban planning objectives—spanning local, regional, and broader contexts. Additionally, developing context-aware architectures will be crucial to better model the spatial relationships between neighboring parcels and ensure greater coherence in generated urban forms.


\newpage
\printbibliography
\appendix
\section{Prompt Design}
\label{app:LLM_prompt}
\noindent Table \ref{tab:comparison_prompt_styles} summarizes some examples of the three prompting styles: minimal, structured, and elaborate prompts. To create the structured prompt, we have summarized its five components as below.

\begin{enumerate}
    \item \textbf{Settlement type}:
    Settlement type is determined using the ``places'' layer in OpenStreetMap, identifying the primary type by area coverage (i.e. city, town, village). 
    Descriptions use the following phrase variation templates:
    \begin{itemize}
        \item ``The area shown in the satellite image of \{city\_name\} falls within a \{type\}"
        \item ``This is a satellite image of \{type\} in \{city\_name\}"
        \item ``This is a satellite image of \{city\_name\} where a \{type\} forms the core"
    \end{itemize}
    When secondary settlement types cover $>35\%$ of the area, they are added using connector phrases:
    \begin{itemize}
        \item ``..., with some \{types\} mixed in"
        \item ``..., alongside portions of \{types\}"
        \item ``..., blending into \{types\} areas"
        \item ``..., adjacent to \{types\} zones"
    \end{itemize}
    
    \item \textbf{Land use composition}: Land use composition represents the proportions of land use categories: residential, commercial, industrial, recreational, farmland, forest, water, and parking. These percentages are calculated using OpenStreetMap - the ``land use'' layer -for all categories except parking, which is derived from the ``traffic'' layer. If any category surpasses a $5\%$ area threshold, one of the following templates is randomly selected to describe the primary land use:
    \begin{itemize}
        \item ``This area is dominated by \{name\} (\{pct\}\%)"
        \item ``The landscape is primarily \{name\} (\{pct\}\%)"
        \item ``\{name\} areas (\{pct\}\%) prevail here"
        \item ``You’ll find mostly \{name\} (\{pct\}\%) in this zone
    \end{itemize}

    Additional land use types are appended with one of these connectors:
    \begin{itemize}
        \item ``..., complemented by \{names (\{pct\}\%)\}”
        \item ``..., with pockets of \{names (\{pct\}\%)\}”
        \item ``..., alongside some \{names (\{pct\}\%)\}”
        \item ``..., interspersed with \{names (\{pct\}\%)\}”
    \end{itemize}

    \item \textbf{Residential type}:
    Within residential areas, the dominant building type—apartment complexes, single-family homes, or townhouses—is described using one of the following:

    \begin{itemize}
        \item ``The residential buildings are mainly \{type\}”
        \item ``Housing consists primarily of \{type\}”
        \item ``\{type\} structures dominate the residential areas”
        \item ``You’ll find mostly \{type\} here”
    \end{itemize}
    
    Additional types are included using:
    \begin{itemize}
        \item ``..., with some \{types\} interspersed”
        \item ``..., complemented by \{types\}”
        \item ``..., alongside \{types\} dwellings”
        \item ``..., mixed with \{types\} residences”
    \end{itemize}

    If residential land is present but no discernible building types are identified, a fallback message is omitted to avoid misrepresentation.
    
    \item \textbf{Building coverage:} 
    
    This prompt complements the land use descriptions by stating the percentage of area occupied by buildings using the building outlines in the OSM ``Building'' layer. At a high level, land use patterns loosely describe the main area functionalities. But it is unknown how much space the buildings occupy, as opposed to roadside infrastructure or facilities. 
    Building density is categorized into high (≥30\%), medium (≥15\%), and low (≥3\%) based on the total building footprint. Density is described using randomly selected templates:
    \begin{itemize}
        \item ``Building density is \{low/medium/high\} in this area”
        \item ``This area has a \{low/medium/high\} building density”
    \end{itemize}

    \item \textbf{(Optional) Landuse designation:}
    In addition to describing land use composition, we introduce explicit spatial cues by referencing shaded land use blocks in the control image. When a land use type occupies a moderate proportion of the tile (10\%–40\%), and is spatially concentrated, we describe its approximate position using the position of its centroid coordinates (horizontal: left/central/right, vertical: lower/mid/upper).
    \begin{itemize}
        \item ``The \{landuse\} area is concentrated in the \{position\} of the image in shaded \{color\}." 
        \item ``A \{landuse\} patch appears in the \{position\} region of the image in shaded \{color\}”
        \item ``\{landuse\} areas cluster in the \{position\} portion of the image in shaded \{color\}”
        \item ``The main \{landuse\} zone is located toward the \{position\} in shaded \{color\}”
    \end{itemize}

\end{enumerate}

\begin{table}[h]
\small
    \centering
    \begin{tabular}{p{0.03\linewidth}|p{0.11\linewidth}|p{0.77\linewidth}}
    \toprule[1pt]
        \# & Style & Prompt \\
        \midrule
        \multirow{3}{*}{1} & Minimal & Satellite image in a city in la. Landuse include: 85\% residential, commercial (10\%). Medium building density. Residential type is mainly single-family homes. \\
        \cmidrule{2-3}
        & Structured & This is a satellite image of city in la. Residential areas (85\%) prevail here, complemented by commercial (10\%). Building density is medium in this area. Housing consists primarily of single-family homes. \\
        \cmidrule{2-3}
        & Elaborate & The provided satellite image depicts an area within the city of La, with land use consisting primarily of residences at 85\%. Commercial areas account for approximately 10\%, which includes various types of businesses such as retail stores, offices, restaurants, etc. The overall building density appears to be moderate, characterized by medium rise structures that are predominantly residential in nature. Residential properties consist mostly of single-family homes, offering spacious living spaces tailored towards families or individuals seeking comfortable suburban lifestyles. \\

        \midrule
        \multirow{3}{*}{2} & Minimal & Satellite image in a city in dallas. Landuse include: 45\% residential, commercial (20\%), forest (15\%). Medium building density. Residential type is mainly apartment complexes, with townhouses. \\
        \cmidrule{2-3}
        & Structured & This is a satellite image of dallas where the city forms the core. Furthermore, residential areas (45\%) prevail here, with pockets of commercial (20\%), forest (15\%). This area has a medium building density. Meanwhile, housing consists primarily of apartment complexes, complemented by townhouses. \\
        \cmidrule{2-3}
        & Elaborate & This satellite image depicts an urban area within the Dallas city limits. The land use distribution comprises of various sectors such as residential areas accounting for approximately 45\%, followed by commercial zones at around 20\%. Forests cover about 15\% of the region's surface. Medium-sized buildings are present throughout the landscape, indicating moderate development levels. Residential types primarily consist of apartment complexes and townhouse communities, providing diverse housing options to residents. \\

        \midrule
        \multirow{3}{*}{3} & Minimal & Satellite image in a city in chicago. Landuse include: 35\% residential, parking (15\%), recreational (10\%), commercial (10\%), forest (5\%). high building density. Residential type is mainly apartment complexes , with single-family homes. \\
        \cmidrule{2-3}
        & Structured & The area shown in the satellite image of chicago falls within the city. You'll find mostly residential (35\%) in this zone, alongside some parking (15\%), recreational (10\%), commercial (10\%), forest (5\%). Building density is high in this area. apartment complexes structures dominate the residential areas, alongside single-family homes dwellings. \\
        \cmidrule{2-3}
        & Elaborate & This satellite view of Chicago shows the distribution of land uses within its borders. The majority of the area consists of residential areas at 35\%, which are primarily composed of apartment complexes for multi-dwelling units; there's also an ample amount of single family homes scattered throughout the landscape. Parking spaces make up approximately 15\% of the total space, providing convenient access to vehicles. Recreational facilities account for around 10\%. Commercial establishments such as shopping centers or offices occupy roughly 10\% of the overall picture. Lastly, forests cover about 5\% of the visible region. In terms of architectural features, it appears that buildings have quite dense concentrations, showcasing the bustling nature of life within these neighborhoods. \\
    \bottomrule[1pt]
    \end{tabular}
    \caption{Three examples of three prompting styles}
    \label{tab:comparison_prompt_styles}
\end{table}

Here is the LLM prompt used for enriching the minimal prompts into a LLM-enriched elaborate prompt. 
\begin{lstlisting}
### Task:
Enrich this satellite image description while:
1. Keeping ALL original numbers/percentages EXACTLY as given, and in numerical form
2. Adding only qualitative details (no new stats)
3. Maintaining professional urban planner tone
4. Be succinct, keep output under 100 words
### Original:
{Description from the minimal version}
### Enriched:
\end{lstlisting}


\newpage
\section*{CRediT authorship contribution statement}
Qingyi Wang: Writing -- review \& editing, Writing -- original draft, Visualization, Validation, Methodology, Investigation, Formal analysis, Data curation. Yuebing Liang: Writing -- review \& editing, Writing -- original draft, Methodology, Investigation. Yunhan Zheng: Writing -- review \& editing. Kaiyuan Xu: Validation, Methodology, Investigation. Jinhua Zhao: Supervision, Project administration. Shenhao Wang: Writing -- review \& editing, Writing -- original draft, Validation, Supervision, Project administration, Methodology, Investigation, Formal analysis, Data curation, Conceptualization.

\end{document}